\documentclass[conference]{IEEEtran}
\IEEEoverridecommandlockouts

\usepackage{url}
\usepackage{cite}
\usepackage{amsmath,amssymb,amsfonts}
\usepackage{accents}
\usepackage[linesnumbered,ruled]{algorithm2e}
\usepackage[noend]{algpseudocode}
\usepackage{graphicx}
\usepackage{textcomp}
\usepackage[table,xcdraw]{xcolor}
\usepackage{xcolor}
\usepackage{comment}
\usepackage{multicol,multirow}
\usepackage[T1]{fontenc}
\def\BibTeX{{\rm B\kern-.05em{\sc i\kern-.025em b}\kern-.08em
    T\kern-.1667em\lower.7ex\hbox{E}\kern-.125emX}}
\usepackage{todonotes}
\usepackage{caption}
\usepackage{wrapfig}
\usepackage{subfig}
\newsavebox{\measurebox}
\usepackage{threeparttable}
\usepackage[font=small]{caption}
\usepackage{enumitem}
\usepackage{booktabs}

\usepackage{hyperref}
\hypersetup{
    colorlinks=false
}

\makeatletter
\newcommand{\algorithmfootnote}[2][\footnotesize]{%
  \let\old@algocf@finish\@algocf@finish
  \def\@algocf@finish{\old@algocf@finish
    \leavevmode\rlap{\begin{minipage}{\linewidth}
    #1#2
    \end{minipage}}%
  }%
}
\makeatother

\begin{document}
\bstctlcite{IEEEexample:BSTcontrol}

\title{
Robustness Testing of Data and Knowledge Driven
Anomaly Detection in Cyber-Physical Systems} 

\author{Regular Paper}

\author{Xugui Zhou, Maxfield Kouzel, Homa Alemzadeh\\

 University of Virginia, Charlottesville, VA 22904, USA \{xz6cz, mak3zaa, ha4d@virginia.edu\}}

\maketitle
\thispagestyle{plain}
\pagestyle{plain}

\vspace{-2em}
\begin{abstract} 
The growing complexity of Cyber-Physical Systems (CPS) and challenges in ensuring safety and security have led to the increasing use of deep learning methods for accurate and scalable anomaly detection. 
However, machine learning (ML) models often suffer from low performance in predicting unexpected data and are vulnerable to accidental or malicious perturbations. 
Although robustness testing of deep learning models has been extensively explored in applications such as image classification and speech recognition, less attention has been paid to ML-driven safety monitoring in CPS. 
This paper presents the preliminary results on evaluating the robustness of ML-based anomaly detection methods in safety-critical CPS against two types of accidental and malicious input perturbations, generated using a Gaussian-based noise model and the Fast Gradient Sign Method (FGSM). We test the hypothesis of whether integrating the domain knowledge (e.g., on unsafe system behavior) with the ML models can improve the robustness of anomaly detection without sacrificing accuracy and transparency. 
Experimental results with two case studies of Artificial Pancreas Systems (APS) for diabetes management show that ML-based safety monitors trained with domain knowledge can reduce on average up to 54.2\% of robustness error and keep the average F1 scores high while improving transparency.

\end{abstract}


\section{Introduction}
\vspace{-0.5em}


Deep learning (DL) methods are increasingly used for anomaly detection~ \cite{luo_deep_2021,dsn2021zhou,ding2021mini} and attack recovery~ \cite{Choi2020software,dash2021piper} in safety-critical CPS, such as medical devices and autonomous vehicles. The ML-based anomaly detection methods are often preferred to model-based techniques due to their easier implementation, powerful capability in capturing the relationship between input and output or approximating dynamic models of the control systems, and high accuracy in predicting unseen data that shares similar features with the training data.

However, the effectiveness of the ML models heavily relies on the quantity and quality of the training data~\cite{sequential2021Md}, especially in safety-critical applications where sufficient and representative data (e.g., anomaly examples) is difficult or expensive to collect. 
This limited data availability impedes the development of accurate ML models for anomaly detection. Accidental and malicious perturbations in the ML input may also cause misclassifications or incorrect predictions~\cite{guo_rnn-test_2021, jia_adversarial_2021} and lead to catastrophic consequences in safety-critical applications~\cite{olowononi_resilient_2021}. Further, the complex architectures of DL networks and their black-box nature lead to a lack of transparency and intractability and make them hard to verify~\cite{ehsan2021expanding,cutillo2020machine}.

Although the robustness of DL models against adversarial perturbations has been extensively studied for computer vision and speech recognition applications, less attention has been paid to the robustness of anomaly detection models against perturbations on multivariate time-series data~\cite{goodfellow2014explaining,kurakin2018adversarial}. Previous works have studied the differences between adversarial machine learning in CPS vs. cyberspace systems \cite{li_conaml_2021} and the challenges of implementing adversarial attacks in CPS \cite{jia_adversarial_2021}. However, robustness testing and possible methods to improve the robustness of safety monitoring have not been well investigated in CPS.

In this paper, we adopt the adversarial example crafting techniques to evaluate the robustness of DL-based safety monitors in CPS. We focus on answering two research questions:
\begin{itemize}
    \item \textbf{RQ1:} How robust are the state-of-the-art ML safety monitors against accidental and malicious input perturbations?
    \item \textbf{RQ2:} Does the integration of domain knowledge help with improving the robustness of ML monitors?
\end{itemize}

\begin{figure*}[]
    \centering
    
    \begin{minipage}{0.85\columnwidth}
	\scriptsize \centering
    \includegraphics[width=\columnwidth]{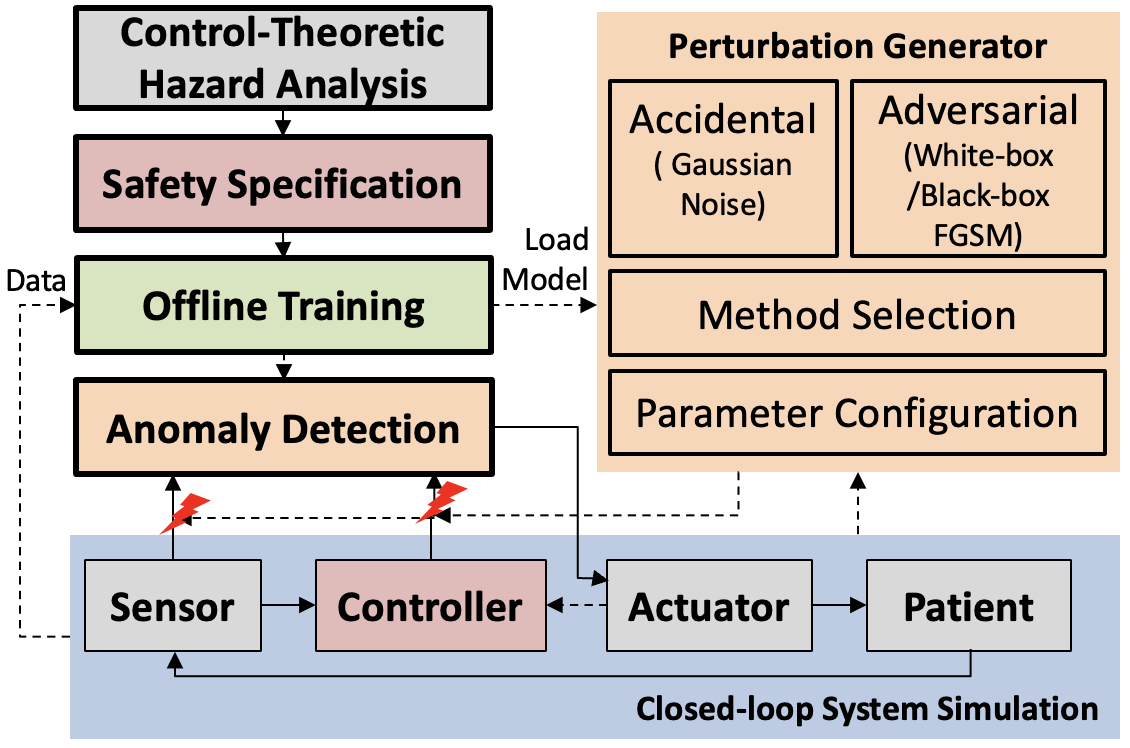}
    
    (a)  
    \end{minipage}
    \begin{minipage}{0.85\columnwidth}\scriptsize \centering
    \includegraphics[width=\columnwidth]{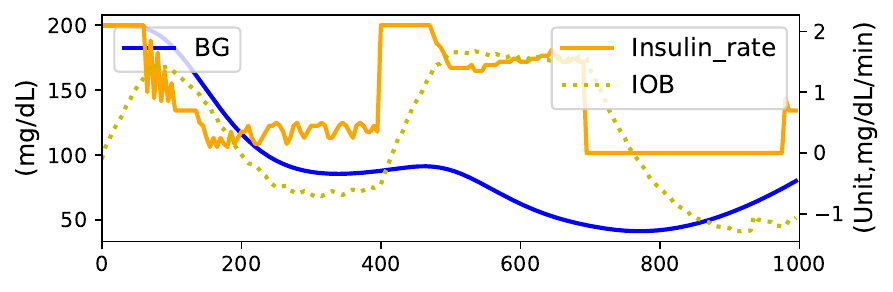}
    \includegraphics[width=\columnwidth]{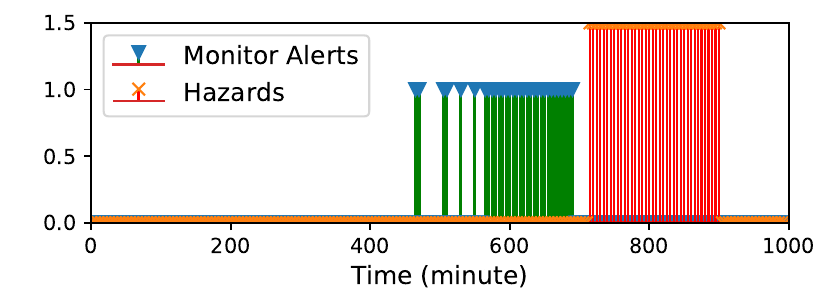}   
     \hfill
    (b)   
    \end{minipage}
    \vspace{-0.5em}
    \caption{(a) Overall Framework. (b) An Example APS Simulation Trace with Safety Monitor.}
    \vspace{-1.5em}
    \label{fig:overallframework}
\end{figure*}

Specifically, we evaluate the robustness of the ML-based safety monitors in detecting unsafe control actions issued by a CPS controller in the presence of accidental Gaussian noise and adversarial perturbations affecting their inputs (sensors and control commands). 
For the integration of domain knowledge with ML models, we extract the context-specific specification of unsafe control actions in a given CPS using a control-theoretic hazard analysis method \cite{leveson2013stpa,dsn2021zhou,zhou2022Strategic} and add the extracted safety properties as a regularization term in a semantic loss function \cite{xu2018semantic,zhou2021knowledge} that guides the training process. Fig. \ref{fig:overallframework} presents the overall framework. 

The main contributions of the paper are as follows:
\begin{itemize}[leftmargin=*]

\item Assessing the resilience of ML-based anomaly detection models against both accidental and malicious perturbations on multivariate time-series input using a Gaussian-based noise model
and the widely-used fast gradient sign method (FGSM) for adversarial example generation. 

\item Integrating domain knowledge into ML monitors as a semantic loss function and evaluating the robustness in comparison to two state-of-the-art baseline monitors, Multi-layer Perceptron (MLP) and Long Short Term Memory (LSTM). 

\item Evaluating the robustness of the data-driven vs. combined data and knowledge driven monitors using datasets collected from two closed-loop Artificial Pancreas Systems (APS) with different controllers and patient glucose simulators. 
Our results show that the ML monitors with semantic loss function can reduce on average up to 54.2\% of robustness error and keep average F1 scores high while improving transparency.


\end{itemize}

\section{Preliminaries}
The core of CPS are controllers that monitor and control physical processes (e.g., patient's dynamics) by estimating the current physical state (e.g., blood glucose (BG) and insulin on board (IOB)) based on sensor measurements and sending control commands (e.g., rate of insulin injection) to the actuators. We consider a safety monitor that is integrated with a CPS controller (e.g., an APS controller) and observes the sensor data and control commands to evaluate whether the control commands issued in a given system context might be unsafe and lead to hazards and to stop their delivery to the actuators (see Fig.~\ref{fig:overallframework})~\cite{dsn2021zhou}. 

\subsection{Machine Learning Based Safety Monitoring}
We model the task of detecting an unsafe control action as a context-specific conditional event, as shown below: 
\begin{equation}
\label{eq:ml-model0}
y_{t} = p(\exists t'\in [t,t+T]: x_{t'} \in {\mathcal{X}}_{h}|f({X}_{t}),f({U}_{t}))
\end{equation}
where, $f(\cdot)$ represents an aggregation function (e.g., average, Euclidean norm, or regression) over a window of sensor measurements ${X}_{t}$ or control actions ${U}_{t}$.  Given the control action sequence ${U}_{t}$ executed under the system state sequence ${X}_{t}$, an ML-based monitor outputs a binary $y_{t}$ that classifies $U_t$ to \textit{safe} or \textit{unsafe} depending on whether a hazard is expected to occur within $T$ timesteps from time $t$. 
Fig.~\ref{fig:overallframework}(b) shows an example simulation trace from monitoring an APS system.  

\subsection{Context-Dependent Safety Specification}
For integrating the domain knowledge on the safety of the control actions issued by a controller, we adopt an approach based on the high-level control-theoretic hazard analysis~{\cite{leveson2013stpa}} for specification of context-dependent safety requirements~{\cite{dsn2021zhou}}. Table \ref{table:stltable} shows an example set of context-dependent safety specifications for the APS case study described as Signal Temporal Logic (STL \cite{Eziospecification}) formulas. Each formula specifies the system context (based on controller inputs and estimated state variables) under which a control action $u_t$ is potentially unsafe and might lead to a safety hazard $H_i$, if issued by the controller. These formulas can be also synthesized into logic to design a rule-based safety monitor solely based on domain knowledge and are applicable to any controller with the same functional specification. 

\begin{table}[b]
\centering
\vspace{-1em}
\caption{Context Dependent Safety Specifications for APS}
\vspace{-0.5em}
\label{table:stltable}
\resizebox{0.9\columnwidth}{!}
{%
\begin{threeparttable}

\begin{tabular}{|c|l|c|} \hline
\textbf{Rule} & \multicolumn{1}{c|}{\multirow{2}{*}{\textbf{STL Description of Safety Context}}} &\textbf{Implied} \\ 
\textbf{$\Phi_{h}$} && \textbf{Hazard Type} \\ \hline

1 & $(BG>BGT\wedge BG'>0)\wedge (IOB'<0)\wedge u_{1}$ & H2 
\\ 
2 & $(BG>BGT\wedge BG'>0)\wedge (IOB'=0)\wedge u_{1}$ & H2 
\\ 
3 & $(BG>BGT\wedge BG'<0)\wedge (IOB'>0)\wedge u_{1}$ & H2 
\\ 
4 & $(BG>BGT\wedge BG'<0)\wedge (IOB'<0)\wedge u_{1}$ & H2 
\\ 
5 & $(BG>BGT\wedge BG'<0)\wedge (IOB'=0)\wedge u_{1}$ & H2 
\\ 
6 & $(BG<BGT\wedge BG'<0)\wedge (IOB'>0)\wedge u_{2}$ & H1 
\\ 
7 & $(BG<BGT\wedge BG'<0)\wedge (IOB'<0)\wedge u_{2}$ & H1 
\\ 
8 & $(BG<BGT\wedge BG'<0)\wedge (IOB'=0)\wedge u_{2}$ & H1 
\\ 
9 & $(BG>BGT)\wedge u_{3}$ & H2
\\ 
10 & $(BG<70)\wedge\neg u_{3}$ & H1 
\\ 
11 & $(BG>BGT\wedge BG'>0)\wedge (IOB'<=0) \wedge u_{4}$ & H2 
\\ 
12 & $(BG<BGT\wedge BG'<0)\wedge (IOB'>=0)\wedge u_{4}$ & H1\\

\hline

\end{tabular}

\begin{tablenotes}\footnotesize
\item[*] BGT: BG target value; 
$BG' =dBG/dt$, $IOB'=dIOB/dt$; 
\item[*] $u_{1,2,3,4}:$decrease\_insulin, increase\_insulin, stop\_insulin, keep\_insulin;
\item[*] $H1:$ Too much insulin is infused, which will reduce the
BG and might lead to hypoglycemia; $H2:$ Too little insulin is infused, which causes the BG to
increase and could lead to hyperglycemia.

\end{tablenotes}

\end{threeparttable}
}
\end{table}
\subsection{Integrating Domain Knowledge with ML}
We encode the STL formulas generated for detecting unsafe control actions as a custom semantic loss function that penalizes the ML model during the training process, if the prediction does not match with any of the unsafe control action formulas. The new loss function is as follows:
\begin{equation}
    \label{eq:loss_custom}
    loss = loss_{ex} + w \left|y_{t}-I\left(\bigvee_{\Phi_h} f(\mu(X_{t}))\models \Phi_{h}\right) \right|
\end{equation}
\noindent where $loss_{ex}$ is the baseline ML model loss function (e.g., cross-entropy loss), $w$ is a weight parameter that determines the degree that system context and safety specification would interfere with the training process, $y_{t}$ is the output prediction of the ML  model, and $I(\cdot)$ is an indicator function indicating whether the aggregated values of the estimated state variables for a measurement window, $f(\mu(X_{t}))$, satisfy any of the unsafe control action specifications $\Phi_{h}$ listed in Table \ref{table:stltable}. 

This approach is generalizable as the safety specifications are generated using a high-level hazard analysis method that is not limited to any specific CPS. In addition, the semantic loss function can be applied to any ML model.




\section{Robustness Testing}
We consider the \textit{accidental or malicious perturbations} on ML-based monitor inputs as small changes that cannot be detected by the current methods for sensor/input error detection and attack detection, such as invariant detection \cite{Adepu2016UsingPI} or change detection techniques (e.g., Cumulative Sum Control Chart (CUSUM)\cite{attack_pcs,jia_adversarial_2021}), but can lead to misclassification results and 
severe consequences for the CPS being monitored. For example, an attacker can remotely login to an insulin pump and change the output control commands \cite{PumpRecall2019}, or due to a malfunction, the pump can deliver an incorrect insulin dosage \cite{PumpRecall2-2019}. If these small perturbations are also sent to the ML-based monitor, either event could evade the detection by the monitor \cite{MedicalCyberattack2019} and result in severe complications such as hypo- or hyperglycemia\cite{marcovecchio2017complications} and potential death. 

\textbf{Simulating Environment Noise on Input Data:}
To generate small accidental perturbations, we add a randomly generated error value from a Gaussian distribution with zero mean and small standard deviation (less than one standard deviation) to the sensor data. More aggressive deviations might be easily detected by the existing CPS anomaly detection techniques~\cite{attack_pcs, Mehmed5794, Adepu2016UsingPI}.

We use Gaussian noise to simulate the environment noise on input data because (1) Gaussian is a reasonable assumption for any process or system that's subject to the Central Limit Theorem \cite{fischer2011history} and (2) Gaussian noise is entirely described by second-order statistics, which are relatively easy to measure. 
Further investigation of the generalization of the proposed methods against a broader range of noises is out of the scope of this paper.

\textbf{White-box Attacks:}
Fast Gradient Sign Method (FGSM) \cite{goodfellow2014explaining} is a simple but effective method widely used in generating adversarial images using the gradients of a neural network, and is reported to be also effective in non-image domain \cite{mode_adversarial_2020}. For a specific input $x$ and loss function $J(\cdot,\cdot)$, the adversarial sample $x^{adv}$ can be generated using the following equations:
\vspace{-0.25em}
\begin{align}
    &x^{adv} = x + \Delta_x \\
    &\Delta_x = \epsilon*sign(\nabla_{x}(J(x,\Bar{y}) )
\end{align}
where $\Delta_x $ is the generated perturbation on the input $x$ with label $\Bar{y}$, and $\epsilon$ is a small constant parameter that limits the strength of perturbations at each dimension. The resulting adversarial input maximizes the loss function using the $L_\infty$ norm \cite{xu2017feature}. Unlike Gaussian noise, which is only applied to sensor data, we inject FGSM attacks to the multivariate time series input data (both sensor and control commands).

\begin{figure}[t]
    \centering
    \includegraphics[width=\columnwidth]{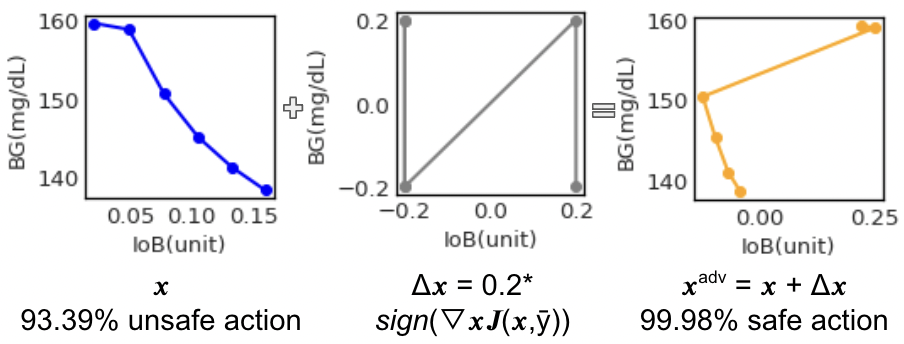}
    \caption{An Example FGSM Attack on a Baseline Monitor with the \textit{Keep\_Insulin} Injection Command Issued by the Controller.
    }
    \label{fig:xadv}
    \vspace{-2em}
\end{figure}

An example of adversarial input generated using FGSM is shown in Fig. \ref{fig:xadv}. Despite only a minute change to the input, the output of the targeted neural network changes from unsafe with 93.39\% confidence to safe with 99.98\% confidence. 

To implement the FGSM attacks, the attacker would need full access to the target ML model, including the model structure and parameters.

\textbf{Black-box Attacks:}
We also test the robustness of ML monitors against black-box attacks~\cite{guo_black-box_2021} where the attacker does not have full access to the ML safety monitor or its model structure. The attacker's capabilities are limited to sending queries to the model and the knowledge of features used by the model.

Previous studies have shown that adversarial examples can transfer across ML models with different architectures \cite{zhou2018transferable, mode_adversarial_2020, mode_crafting_2020}. So for a black-box attack, the attacker can first train a substitute model using the input/output data from the target safety monitor, then generate white-box adversarial perturbations based on the substitute model, expecting that they are transferable to the target model. 

We use a two-layer MLP (128-64) as the substitute model and generate the adversarial perturbations using the same FGSM approach. Similar to white-box FGSM attacks, the black-box FGSM attacks also target the multivariate input data. 


\section{Experimental Setup and Results}
We used an open-source simulation environment that integrates the closed-loop simulation of two example APS control systems~\cite{dsn2021zhou} to evaluate different ML-based safety monitors. 
Specifically, it integrates two widely-used APS controllers (OpenAPS \cite{openSourceOpenAPS} and Basal-Bolus \cite{BBcontroller}) with two different patient glucose simulators, including Glucosym~\cite{openSourceGlucosym} and UVA-Padova Type 1 Diabetes Simulator~\cite{man2014uva}, simulating 20 different diabetic patient profiles.  

We ran the experiments on an x86\_64 PC with an Intel Core i9 CPU @ 3.50GHz and 32GB RAM running Linux Ubuntu LTS. We used TensorFlow v.2.5.0 to train our ML models.

\subsection{Baseline ML Monitors}

We evaluated two ML architectures, the Multi-layer Perceptron (MLP) and Long Short Term Memory (LSTM). The MLP monitor consisted of two fully connected layers, comprising 256 and 128 neutrons, followed by a fully-connected layer with ReLU activation and a final softmax layer to obtain the hazard probabilities. 
The LSTM monitor was a two-layer (128-64) stacked LSTM with an input time step of 6 (i.e., 30 minutes of data), followed by a fully-connected layer with softmax activation. 
We trained both models using the Adam optimizer with a sparse categorical cross-entropy loss function and a default learning rate of 0.001.




\subsection{Metrics}
We use the following metrics for our experiments:

\begin{table}[b!]
\centering
\vspace{-1em}
\caption{Confusion Matrix for Sequential Data with Tolerance Window $\delta$}
\label{tab:confusionmatrix}

\vspace{-0.5em}

\resizebox{\columnwidth}{!}{%
\begin{threeparttable}

\begin{tabular}{|l|l|l|}
\hline
 & Ground Truth Positive & Ground Truth Negative \\ \hline
PP &  $\sum_{t'=t-\delta'_{t}}^{t}P(t')>0$ \ \&\& $\sum_{t'=t}^{t+\delta}G(t')>0$ &  $P(t)>0$ \&\& $\sum_{t'=t}^{t+\delta}G(t')==0$\\\hline

PN & $\sum_{t'=t-\delta'_{t}}^{t}P(t')==0$ \ \&\& $\sum_{t'=t}^{t+\delta}G(t')>0$  & $P(t)==0$ \&\& $\sum_{t'=t}^{t+\delta}G(t')==0$\\\hline

\end{tabular}%

\begin{tablenotes}\footnotesize
\item[*] PP: Predicted positive; PN: Predicted negative; $P(t)$/$G(t)$: Prediction/Ground truth at time $t$; 
$t-{\delta}'_{t}$: Start time of a window $\delta$, ending with a positive ground truth, that includes t.
\end{tablenotes}

\end{threeparttable}

}
\end{table}
\textbf{{Prediction Accuracy}} represents the performance of the ML-based safety monitors in accurate prediction of hazards, measured using precision, recall, accuracy (ACC), and F1 score calculated using a \textit{Sample Level with Tolerance Window} metric \cite{dsn2021zhou}, where a tolerance window before the start time of hazard ($t_h$) is used for calculation of the metrics. Table \ref{tab:confusionmatrix} shows the confusion matrix with a tolerance window.

\textbf{Prediction Robustness Error} measures how robust the ML model predictions are against the perturbations and is defined as the number of samples that fail to keep the same predicted output classes after adding perturbations, over the total number of samples of each class in the dataset ~\cite{zhang2018cost,basak_universal_2021}:
\begin{equation}    
    \label{eq:robustness_error}
    robustness\ error = \frac{\sum_{i}^{N}I(f_{\theta} (x_i)\neq f_{\theta} (x_i+\Delta_{x} ))}{\sum_{j}^{}N_j}
\end{equation}

where $N_j,N$ is the total number of samples in class $j$ and all the classes, respectively, and $\Delta_x$ is the input perturbation on the ML model $\theta$. The indicator function $I(\cdot)$ equals one when the prediction $f_{\theta}(x_i)$ is not the same as $f_{\theta}(x_i+\Delta_x$).

\subsection{Performance of ML based Safety Monitors}

\begin{table}[t!]
\caption{Overall Performance of Each ML Models without Noises.}
\label{tab:overallperformance}
\resizebox{\columnwidth}{!}{%
\begin{tabular}{|l|l|c|c|c|c|}
\hline
\textbf{Simulator}                                                                         & \textbf{Model} & \textbf{No. Sim.} & \textbf{No. Sample} & \textbf{ACC} & \textbf{F1} \\ \hline
\multirow{5}{*}{\textbf{\begin{tabular}[c]{@{}l@{}}Glucosym\\ (OpenAPS)\end{tabular}}}     & Rule-based           & 8800              & 1.32E+06            & 0.87         & 0.73        \\ \cline{2-6} 
                                                                                           & MLP            & 8800              & 1.32E+06            & 0.97         & 0.89        \\ \cline{2-6} 
                                                                                           & LSTM           & 8800              & 1.32E+06            & 0.98         & 0.93        \\ \cline{2-6} 
                                                                                           & MLP-Custom     & 8800              & 1.32E+06            & 0.98         & 0.91        \\ \cline{2-6} 
                                                                                           & LSTM-Custom    & 8800              & 1.32E+06            & 0.97         & 0.86        \\ \hline
\multirow{5}{*}{\textbf{\begin{tabular}[c]{@{}l@{}}T1DS2013\\ (Basal-Bolus)\end{tabular}}} & Rule-based           & 8800              & 1.32E+06            & 0.61         & 0.56        \\ \cline{2-6} 
                                                                                           & MLP            & 8800              & 1.32E+06            & 0.94         & 0.71        \\ \cline{2-6} 
                                                                                           & LSTM           & 8800              & 1.32E+06            & 0.99         & 0.95        \\ \cline{2-6} 
                                                                                           & MLP-Custom     & 8800              & 1.32E+06            & 0.96         & 0.82        \\ \cline{2-6} 
                                                                                           & LSTM-Custom    & 8800              & 1.32E+06            & 0.98         & 0.90        \\ \hline
\end{tabular}%

}
\vspace{-1em}
\end{table}

\begin{figure}[b]
    \centering
    \vspace{-1.5em}
    \includegraphics[width=\columnwidth]{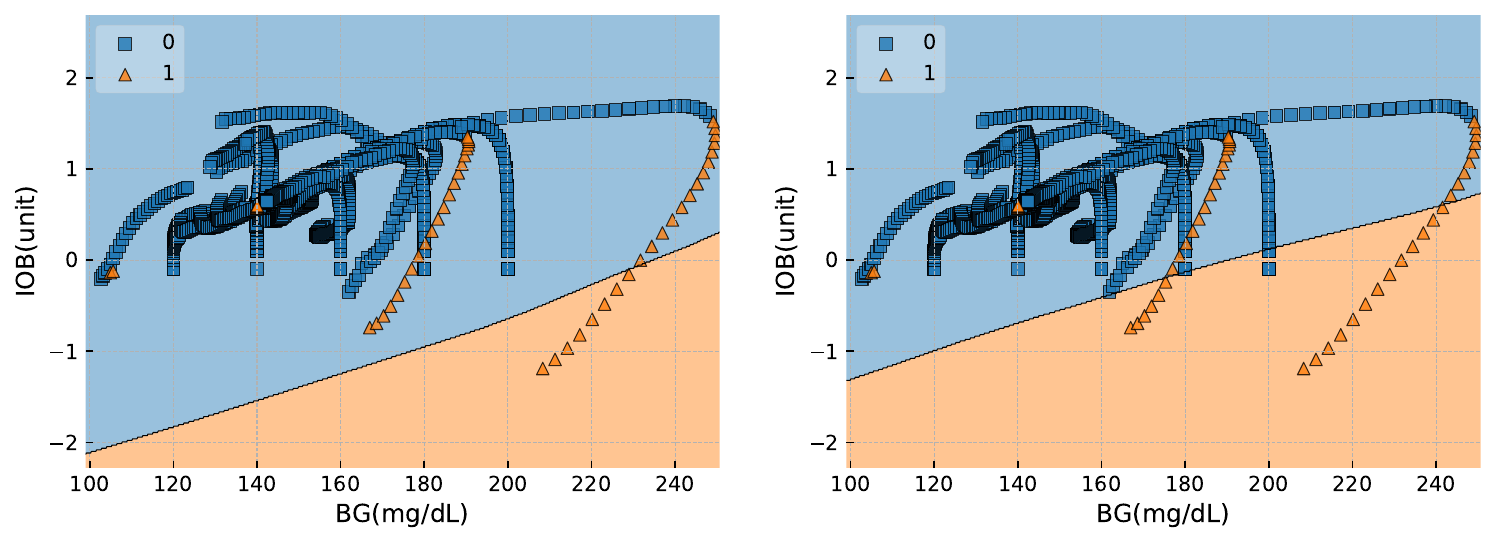}
    \caption{An Example of Decision Boundaries of the Baseline MLP Monitor (Left) and the MLP-Custom Monitor (Right). }
    \label{fig:mlpBoundary}
\end{figure}

Table. \ref{tab:overallperformance} presents the overall performance of the baseline Rule-based and ML monitors (MLP and LSTM) and monitors customized with a semantic loss function based on domain knowledge (MLP-Custom and LSTM-Custom) in detecting anomalies in both simulators in the absence of any perturbations. We see that MLP monitors trained with the custom loss function achieved higher F1 scores than both baseline MLP monitors and the pure Rule-based monitor, indicating the advantage of combining domain knowledge and data-driven techniques. Fig. \ref{fig:mlpBoundary} shows an example of the decision boundaries learned by the MLP and MLP-Custom monitors. Although the LSTM models trained with the custom loss function did not improve the F1 score of baseline LSTM monitors, they achieved comparable accuracy.

Besides, the integration of domain knowledge also improves the ML explainability by offering simple rules to check the output of the ML model.

\subsection{Robustness against Gaussian Noise}

\begin{figure}[t!]
    \vspace{-1.5em}
    \centering
    
    
    
    \includegraphics[width=0.8\columnwidth]{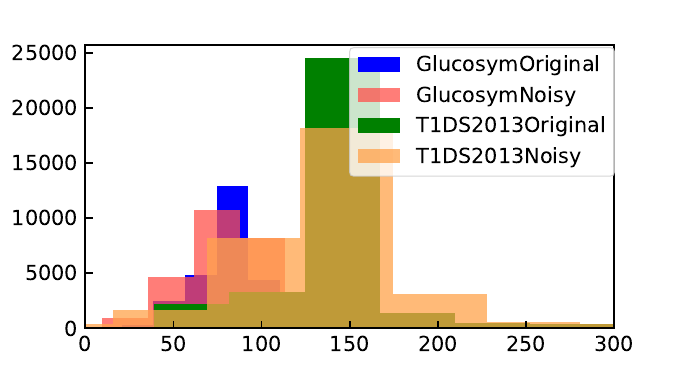}
    \caption{Example Distributions of Test Dataset with/without Adding Gaussian Noise $\mathcal{N}(\mu=0,\sigma^2=(0.5std)^2)$.}
    \vspace{-1em}
    \label{fig:distribution-Gaussian}
\end{figure}

An example distribution of the input dataset with/without Gaussian noises is presented in Fig. \ref{fig:distribution-Gaussian}.

\begin{figure}[b!]
    \vspace{-1.5em}
    \centering
    
    
    
    \begin{minipage}{\columnwidth}
	\scriptsize \centering
    \includegraphics[width=0.9\columnwidth]{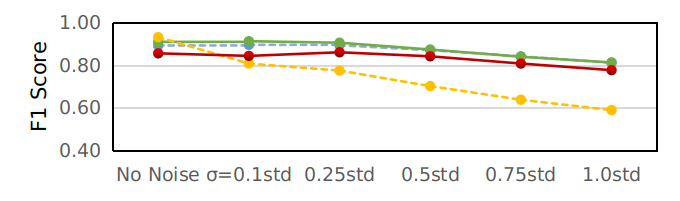}
    
    (a)  Glucosym Simulator with OpenAPS Controller.
    \end{minipage}
    \hfill
    \begin{minipage}{\columnwidth}\scriptsize \centering
    \includegraphics[width=0.9\columnwidth]{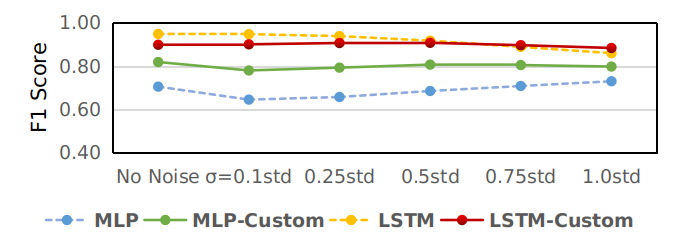}
    
    (b)   T1DS2013 Simulator with Basal-Bolus Controller.
    \end{minipage}
    
    \caption{F1 Score of the ML Models in presence of Gaussian Noise $\mathcal{N}(\mu=0,\sigma^2)$.
    }
    \label{fig:f1-gaussian}
\end{figure}

Fig. \ref{fig:f1-gaussian} shows the performance of the ML models (averaged F1 Score over all the patient profiles) in presence of different levels of input noise ($\sigma=[0.1,0.25,0.5,0.75,1.0]std$) with both Glucosym and T1DS2013 simulators.

We can see from Fig. \ref{fig:f1-gaussian} that both MLP and LSTM models' performance decreases after adding Gaussian noises in both simulators. 
However, for the Glucosym simulator, the LSTM monitor has a more significant drop in F1 score than the MLP monitor (36.8\% vs. 9.1\%), indicating that the LSTM model is less robust against small input perturbations. 
For the T1DS2013 simulator, the LSTM monitor has less drop in F1 score due to the difference in sensor data distribution (see Fig. \ref{fig:distribution-Gaussian}) and the higher percentage of faulty samples (39.3\% vs. 33.9 \%).

\begin{figure}[t!]
    \centering
    \includegraphics[width=0.9\columnwidth]{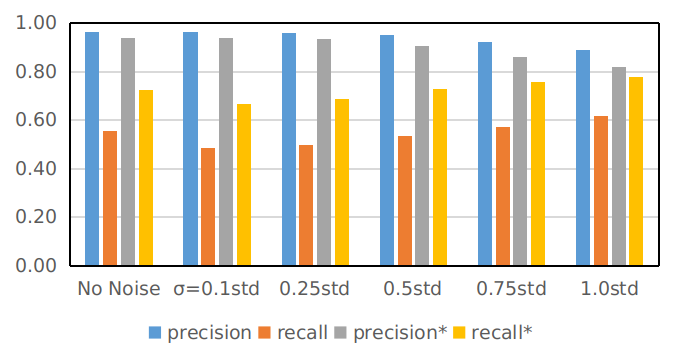}
    \vspace{-0.5em}
    \caption{Precession and Recall of MLP Monitors in T1DS2013 Simulator Against Gaussian Noise $\mathcal{N}(\mu=0,\sigma^2)$. The MLP model retrained with the custom loss function is marked with *.}
    \label{fig:precession-recall}
    \vspace{-1em}
\end{figure}

We also observe that the F1 score of the MLP monitor increases with the rising levels of added noise (higher deviations). This might be because of the large number of new alarms generated due to noisy input that increased the initially low recall of baseline MLP monitor and also resulted in decreased precision, as shown in Fig. \ref{fig:precession-recall}.

In both simulators, the monitors retrained with custom semantic loss functions reduced the performance drop and kept the F1 score high with noisy input data. This demonstrates the advantage of the integrated domain knowledge in overcoming disturbance in the input data as well as the robustness and generalizability in maintaining stable performance for anomaly detection with different ML models and different controllers. 

\subsection{Robustness against White-box FGSM Attacks}

\begin{figure}[b!]
    \vspace{-1em}
    \centering
	\begin{minipage}{0.48\columnwidth}
	\scriptsize \centering
    \includegraphics[width=\columnwidth]{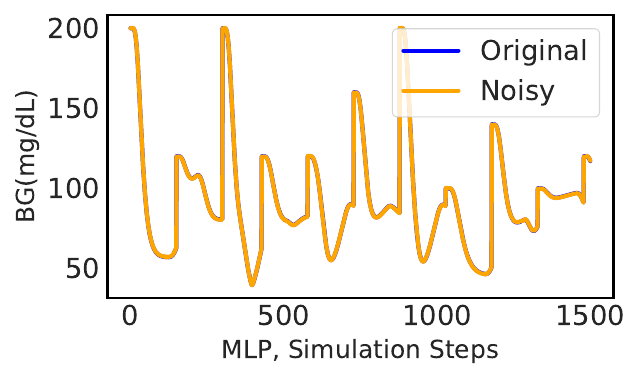}
    
    \end{minipage}
    \hfill
    \begin{minipage}{0.48\columnwidth}\scriptsize \centering
    \includegraphics[width=\columnwidth]{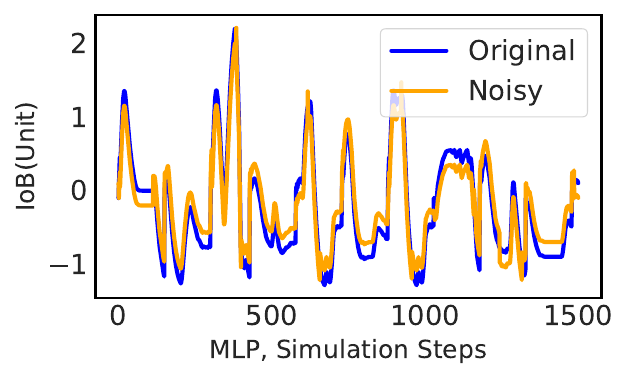}
    
    \end{minipage}
    \hfill
    
    \begin{minipage}{0.48\columnwidth}\scriptsize \centering
    \includegraphics[width=\columnwidth]{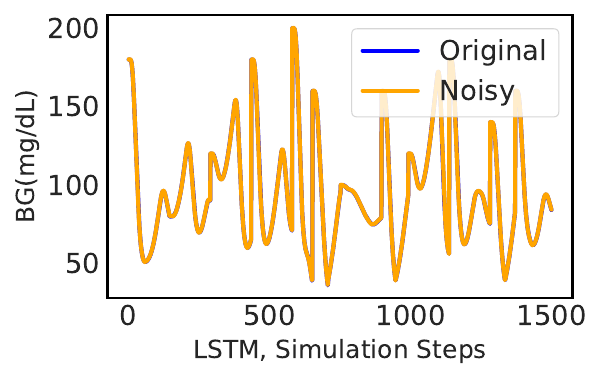}
    
    \end{minipage}
    \hfill
    \begin{minipage}{0.48\columnwidth}\scriptsize \centering
    \includegraphics[width=\columnwidth]{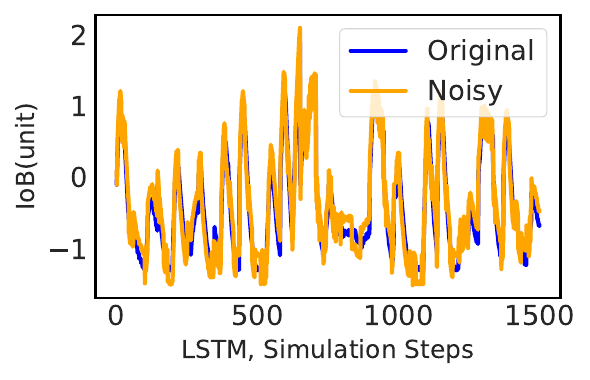}
    
    \end{minipage}
    \hfill
    
    \vspace{-0.5em}	
    \caption{Example Input Data of the MLP and LSTM Models with/without White-box FGSM Attacks ($\epsilon=0.2$). Each simulation step equals 5 minutes in the actual system.}
    \label{fig:signal-FGSM}
\end{figure}

\begin{figure}[t!]
    \centering
    
    
    \begin{minipage}{\columnwidth}
	\scriptsize \centering
    \includegraphics[width=0.9\columnwidth]{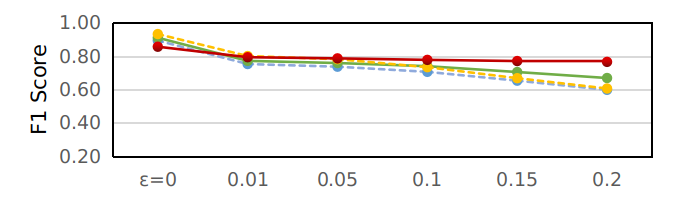}
    
    (a)  Glucosym Simulator with OpenAPS Controller.
    \end{minipage}
    \hfill
    \begin{minipage}{\columnwidth}\scriptsize \centering
    \includegraphics[width=0.9\columnwidth]{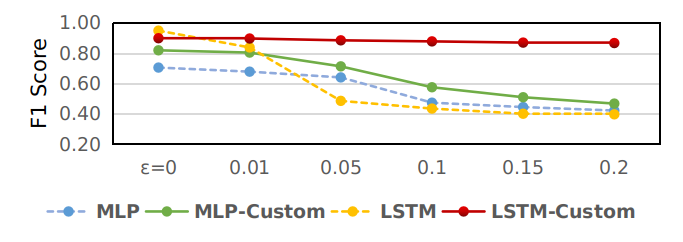}
    
    (b)   T1DS2013 Simulator with Basal-Bolus Controller.
    \end{minipage}
    
    \vspace{-0.5em}	
    \caption{F1 Score of each ML Models Against FGSM Attacks.
    \vspace{-1.5em}
    }
    \label{fig:f1-FGSM}
\end{figure}

An example of adversarial inputs generated by the white-box FGSM attack is shown in Fig. \ref{fig:signal-FGSM}.
Fig. \ref{fig:f1-FGSM} presents the overall F1 score of each ML model (averaged over all the patients) in the presence of white-box FGSM attacks with the adversarial degree $\epsilon$ ranging from 0.01 to 0.2. 
We observe that the F1 scores for both the MLP and LSTM baseline monitors drop with adversarial inputs in both simulators, indicating these ML models' incapability to keep stable performance under adversarial attacks as well as the effectiveness of white-box FGSM attacks in fooling ML models for anomaly detection. 

We also observe that the baseline LSTM monitor has a relatively larger decrease in F1 score than the baseline MLP monitor in the T1DS2013 simulator, which might be because of the differences in the generated adversarial signals (see Fig. \ref{fig:signal-FGSM}) and in the neural network architectures between the MLP and LSTM models. FGSM attacks may have less influence on the MLP models since MLPs are model networks with no memory capacity \cite{DONG20217675}, and the perturbation does not accumulate over a sequence of input data.

After retraining the baseline monitors with the proposed custom loss function, we can observe the improvements of both the MLP monitor and the LSTM monitor in all simulators, demonstrating the effectiveness of the custom loss function in improving ML models' robustness against adversarial attacks. 
Besides, the LSTM-Custom monitors have higher F1 scores than MLP-Custom monitors due to their advantage in dealing with time-series data \cite{lstm2015lipton}.
Therefore, by combining the advantage of integrating domain knowledge and the benefit of the LSTM model structure, the LSTM-Custom monitors maintained the highest F1 scores for both APS systems.

\begin{figure*}[]
    \vspace{-1.5em}
    \centering
	\begin{minipage}{\columnwidth}
	\scriptsize \centering
    \includegraphics[width=0.95\columnwidth]{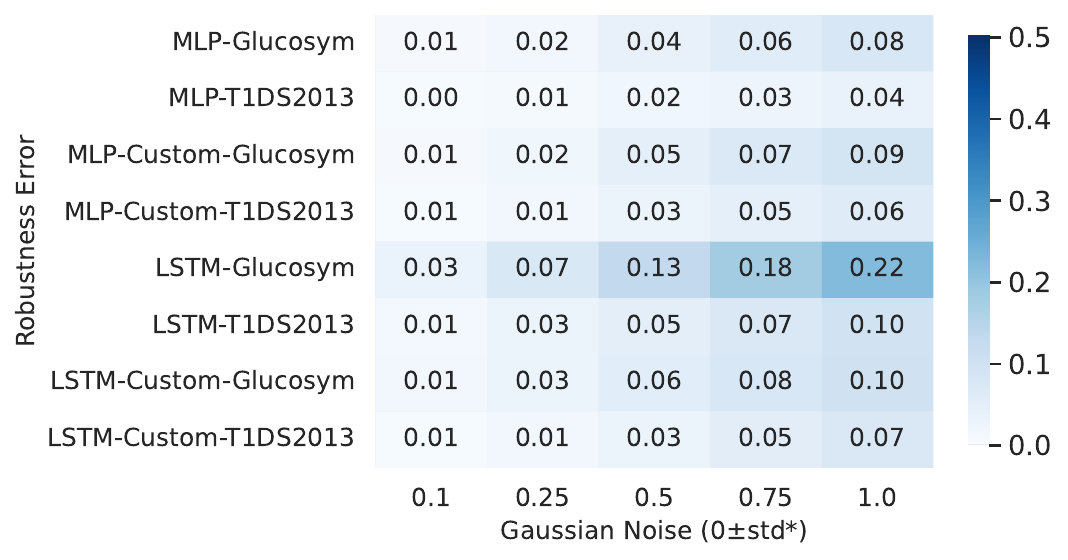}
    
    \end{minipage}
    \hfill
    \begin{minipage}{0.95\columnwidth}\scriptsize \centering
    \includegraphics[width=\columnwidth]{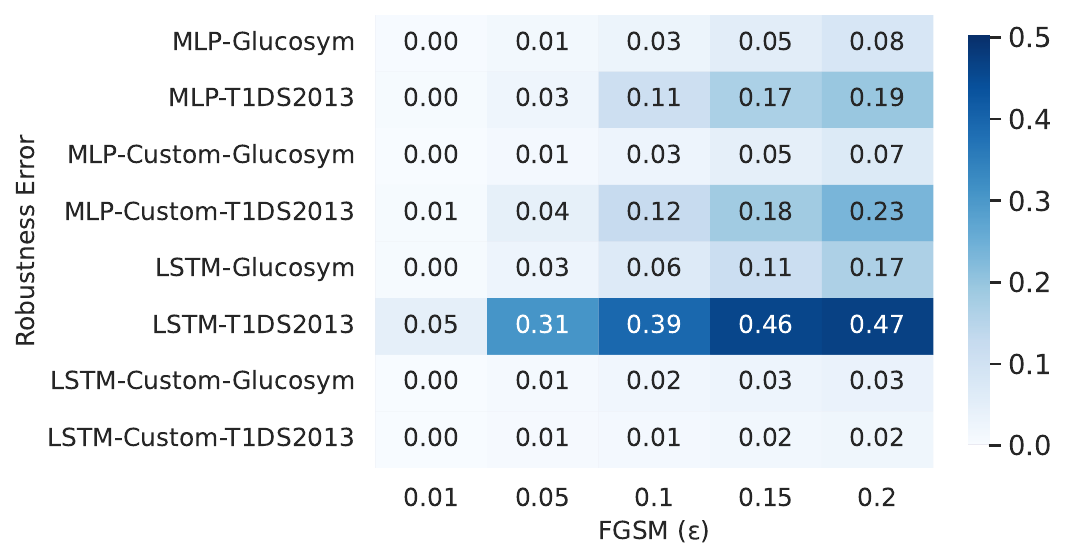}
    
    \end{minipage}
    
    \vspace{-0.5em}

    \caption{Robustness Error of ML Monitors Against Gaussian Noise (Zero Mean and 0.1-1 Standard Deviation) and FGSM Attacks.
    }
    \vspace{-1.5em}
    \label{fig:heatmap-robust}
\end{figure*}

\subsection{Robustness Error Evaluation}
We also evaluate the robustness of the ML monitors by comparing their outputs using the robustness error metric (Eq. \ref{eq:robustness_error}). Fig. \ref{fig:heatmap-robust} shows the heat-map of each ML model's robustness error against the Gaussian noise and FGSM attacks. 

We make the following observations: 
\begin{itemize}
    \item Baseline ML monitors are more vulnerable to FGSM attacks (with larger robustness errors) than simulated environment noise in the input data in both APS simulators.
    \item The baseline LSTM monitor is more sensitive to both FGSM attacks and Gaussian noise due to the disturbance on both their current sensor data and short memory~\cite{DONG20217675,lstm2015lipton}.
    \item Almost all the ML models have more significant robustness error against FGSM attacks in the T1DS2013 simulator than in the Glucosym simulator, which might result from different patient profiles and the more straightforward controller (Basal-Bolus \cite{BBcontroller}) used with the T1DS2013 simulator.
    \item The ML models customized with semantic loss functions have the least robustness error almost in all the situations and reduce on average up to 22.2\% and 54.2\% of the robustness error against Gaussian noise and FGSM attacks across models and simulators. This further attests the advantage of the integrating domain knowledge in keeping the model predictions stable and robust.
\end{itemize}



\subsection{Robustness against Black-box Attacks}

\begin{figure}[t]
    \centering
    \includegraphics[width=0.95\columnwidth]{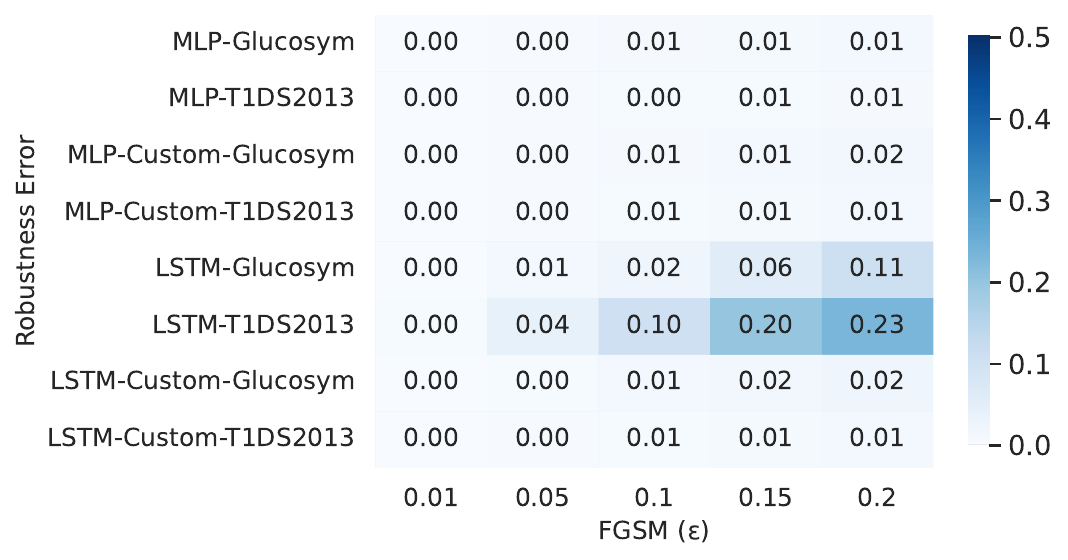}
    \vspace{-0.5em}
    \caption{Robustness Error of ML Monitors Against Black-box Attacks}
    \vspace{-1.5em}
    \label{fig:blackbox-robust}
\end{figure}

We evaluate the robustness of the ML monitors against black-box attacks generated using a substitute MLP model. 
Fig. \ref{fig:blackbox-robust} shows the robustness error heat-map for the ML monitors against the black-box FGSM attacks.
We see that the baseline LSTM models keep the robustness error below 0.23, which is 2.04 times less than the robustness error against white-box FGSM attacks. On the other hand, the baseline MLP models and customized ML models keep the robustness error very small. The inclusion of the custom loss function reduced the robustness error to less than 10\% of its original value.

\section{Related Work}

\textbf{DNN-based anomaly detection in CPS:}
Previous efforts on anomaly detection using Deep Neural Networks have achieved considerable accuracy with well-tuned parameters or complex model structures\cite{luo_deep_2021, yasar_dsn2020, Choi2020software, moon_ensemble_2022, rigaki_survey_2021-1} that suffer from a lack of generalization and transparency. In this paper, we show that the integration of the domain knowledge using a semantic loss function into the anomaly detection model can offer a way to verify the model predictions and improve transparency while keeping the accuracy high.

\textbf{Adversarial attack methods:} Generating adversarial examples to test the robustness of classifiers has been an active area of research
\cite{goodfellow2014explaining,kurakin2018adversarial}. However, most previous works focused on the image classification and speech recognition domains. Few recent works have adopted these advanced attack methods in the non-image domain, such as testing regression models with finance and power consumption data~\cite{mode_adversarial_2020} \cite{mode_crafting_2020}.
There are also some works looking into attacks specifically for CPS, such as testing a universal adversarial method \cite{basak_universal_2021} on the NASA turbofan dataset or crafting adversarial examples for CPS with sensor constraints \cite{li_conaml_2021}. Nevertheless, to the best of the authors' knowledge, this paper is the first to study the impact of adversarial attacks on multivariate time series anomaly detection in CPS.

\textbf{Adversarial defense methods:}
The adversarial training using the adversarial perturbations on input data is one of the most commonly used methods to defend ML models against adversarial examples~\cite{tramer2017ensemble,ganin2016domain,wong2020fast}. However, it suffers from a high cost to generate adversarial examples, and it is impossible to cover all possible adversary techniques that attackers might utilize~\cite{towards2016nicolas}. 
Gradient masking is another common technique for reducing the ML models' sensitivity to small input perturbations, which works by adding a penalty term to the original prediction loss to produce near-zero gradients~\cite{rifai2011contractive,gu2014towards}. However, previous works have shown that this approach scarifies the ML model accuracy~\cite{towards2016nicolas}.
Our work 
deviates from these works by integrating domain knowledge with the ML model as a semantic loss that can improve the ML monitor's robustness and transparency while keeping the accuracy and F1 score high.


\vspace{-0.5em}

\section{Conclusion}
\label{sec:conclusion}

In this paper, we adapt two strategies to craft accidental and adversarial perturbations on multivariate time series data used for anomaly detection in CPS. We also propose a method to protect ML safety monitors against such perturbations by integrating domain knowledge through customizing ML models with semantic loss functions.
We evaluate the robustness of two different ML monitors with and without the proposed protection using the data collected from two closed-loop APS systems. Robustness testing of ML monitors showed large reductions in model performance, despite the models performing very well on unaltered data. The incorporation of domain knowledge significantly improved robustness when compared with baseline models without sacrificing accuracy or transparency. Experimental results showed that baseline LSTM models were more susceptible to attacks than the MLP models, but the LSTM models customized with semantic loss had the least robustness error against all perturbations. These findings warrant a more comprehensive investigation of robustness testing and defense strategies for ML anomaly detection models in CPS.



\section*{Acknowledgment}
 This work was partially supported by the Commonwealth of Virginia under Grant CoVA CCI: C-Q122-WM-02 and by the National Science Foundation (NSF) under Grant No. 1748737.

\bibliographystyle{IEEEtran}
\bibliography{main}

\begin{thebibliography}{10}
\providecommand{\url}[1]{#1}
\csname url@samestyle\endcsname
\providecommand{\newblock}{\relax}
\providecommand{\bibinfo}[2]{#2}
\providecommand{\BIBentrySTDinterwordspacing}{\spaceskip=0pt\relax}
\providecommand{\BIBentryALTinterwordstretchfactor}{4}
\providecommand{\BIBentryALTinterwordspacing}{\spaceskip=\fontdimen2\font plus
\BIBentryALTinterwordstretchfactor\fontdimen3\font minus
  \fontdimen4\font\relax}
\providecommand{\BIBforeignlanguage}[2]{{%
\expandafter\ifx\csname l@#1\endcsname\relax
\typeout{** WARNING: IEEEtran.bst: No hyphenation pattern has been}%
\typeout{** loaded for the language `#1'. Using the pattern for}%
\typeout{** the default language instead.}%
\else
\language=\csname l@#1\endcsname
\fi
#2}}
\providecommand{\BIBdecl}{\relax}
\BIBdecl

\bibitem{luo_deep_2021}
Y.~Luo, Y.~Xiao, L.~Cheng \emph{et~al.}, ``Deep {Learning}-based {Anomaly}
  {Detection} in {Cyber}-physical {Systems}: {Progress} and {Opportunities},''
  \emph{ACM Computing Surveys}, vol.~54, no.~5, pp. 106:1--106:36, May 2021.

\bibitem{dsn2021zhou}
X.~Zhou, B.~Ahmed, J.~H. Aylor \emph{et~al.}, ``Data-driven design of
  context-aware monitors for hazard prediction in artificial pancreas
  systems,'' in \emph{2021 51st Annual IEEE/IFIP International Conference on
  Dependable Systems and Networks (DSN)}, 2021, pp. 484--496.

\bibitem{ding2021mini}
A.~Ding, P.~Murthy, L.~Garcia \emph{et~al.}, ``Mini-me, you complete me!
  data-driven drone security via dnn-based approximate computing,'' in
  \emph{24th International Symposium on Research in Attacks, Intrusions and
  Defenses}, 2021, pp. 428--441.

\bibitem{Choi2020software}
H.~Choi, S.~Kate, Y.~Aafer \emph{et~al.}, ``Software-based realtime recovery
  from sensor attacks on robotic vehicles,'' in \emph{23rd International
  Symposium on Research in Attacks, Intrusions and Defenses ({RAID}
  2020)}.\hskip 1em plus 0.5em minus 0.4em\relax {USENIX} Association, Oct.
  2020, pp. 349--364.

\bibitem{dash2021piper}
P.~Dash, G.~Li, Z.~Chen \emph{et~al.}, ``Pid-piper: Recovering robotic vehicles
  from physical attacks,'' in \emph{51st Annual IEEE/IFIP International
  Conference on Dependable Systems and Networks (DSN)}, 2021, pp. 26--38.

\bibitem{sequential2021Md}
M.~M. Rahman, R.~M. Voyles, J.~Wachs \emph{et~al.}, ``Sequential prediction
  with logic constraints for surgical robotic activity recognition,'' in
  \emph{2021 30th IEEE International Conference on Robot Human Interactive
  Communication (RO-MAN)}, 2021, pp. 468--475.

\bibitem{guo_rnn-test_2021}
J.~Guo, Q.~Zhang, Y.~Zhao \emph{et~al.}, ``{RNN}-{Test}: {Towards}
  {Adversarial} {Testing} for {Recurrent} {Neural} {Network} {Systems},''
  \emph{IEEE Transactions on Software Engineering}, pp. 1--1, 2021.

\bibitem{jia_adversarial_2021}
\BIBentryALTinterwordspacing
Y.~Jia, J.~Wang, C.~M. Poskitt \emph{et~al.},
  ``\BIBforeignlanguage{en}{Adversarial attacks and mitigation for anomaly
  detectors of cyber-physical systems},''
  \emph{\BIBforeignlanguage{en}{International Journal of Critical
  Infrastructure Protection}}, vol.~34, p. 100452, Sep. 2021. [Online].
  Available:
  \url{https://www.sciencedirect.com/science/article/pii/S1874548221000445}
\BIBentrySTDinterwordspacing

\bibitem{olowononi_resilient_2021}
F.~O. Olowononi, D.~B. Rawat, and C.~Liu, ``Resilient {Machine} {Learning} for
  {Networked} {Cyber} {Physical} {Systems}: {A} {Survey} for {Machine}
  {Learning} {Security} to {Securing} {Machine} {Learning} for {CPS},''
  \emph{IEEE Communications Surveys Tutorials}, vol.~23, no.~1, pp. 524--552,
  2021, conference Name: IEEE Communications Surveys Tutorials.

\bibitem{ehsan2021expanding}
U.~Ehsan, Q.~V. Liao, M.~Muller \emph{et~al.}, ``Expanding explainability:
  towards social transparency in ai systems,'' in \emph{Proceedings of the 2021
  CHI Conference on Human Factors in Computing Systems}, 2021, pp. 1--19.

\bibitem{cutillo2020machine}
C.~M. Cutillo, K.~R. Sharma, L.~Foschini \emph{et~al.}, ``Machine intelligence
  in healthcare—perspectives on trustworthiness, explainability, usability,
  and transparency,'' \emph{NPJ digital medicine}, vol.~3, no.~1, pp. 1--5,
  2020.

\bibitem{goodfellow2014explaining}
I.~Goodfellow, J.~Shlens, and C.~Szegedy, ``Explaining and harnessing
  adversarial examples,'' \emph{arXiv 1412.6572}, Dec. 2014.

\bibitem{kurakin2018adversarial}
A.~Kurakin, I.~J. Goodfellow, and S.~Bengio, ``Adversarial examples in the
  physical world,'' in \emph{Artificial intelligence safety and
  security}.\hskip 1em plus 0.5em minus 0.4em\relax Chapman and Hall/CRC, 2018,
  pp. 99--112.

\bibitem{li_conaml_2021}
\BIBentryALTinterwordspacing
J.~Li, Y.~Yang, J.~S. Sun \emph{et~al.}, ``\BIBforeignlanguage{en}{{ConAML}:
  {Constrained} {Adversarial} {Machine} {Learning} for {Cyber}-{Physical}
  {Systems}},'' in \emph{\BIBforeignlanguage{en}{Proceedings of the 2021 {ACM}
  {Asia} {Conference} on {Computer} and {Communications} {Security}}}.\hskip
  1em plus 0.5em minus 0.4em\relax Virtual Event Hong Kong: ACM, May 2021, pp.
  52--66. [Online]. Available:
  \url{https://dl.acm.org/doi/10.1145/3433210.3437513}
\BIBentrySTDinterwordspacing

\bibitem{leveson2013stpa}
N.~Leveson and J.~Thomas, ``An {STPA} primer,'' \emph{Cambridge, MA}, 2013.

\bibitem{zhou2022Strategic}
X.~Zhou, A.~Schmedding, H.~Ren \emph{et~al.}, ``Strategic safety-critical
  attacks against an advanced driver assistance system,'' in \emph{2022 52nd
  Annual IEEE/IFIP International Conference on Dependable Systems and Networks
  (DSN)}, 2022.

\bibitem{xu2018semantic}
J.~Xu, Z.~Zhang, T.~Friedman \emph{et~al.}, ``A semantic loss function for deep
  learning with symbolic knowledge,'' in \emph{International conference on
  machine learning}.\hskip 1em plus 0.5em minus 0.4em\relax PMLR, 2018, pp.
  5502--5511.

\bibitem{zhou2021knowledge}
X.~Zhou, B.~Ahmed, J.~H. Aylor \emph{et~al.}, ``Knowledge and data driven
  synthesis of runtime monitors for cyber-physical systems,'' in \emph{under
  review of IEEE Transactions on Dependable and Secure Computing (TDSC)}, 2021.

\bibitem{Eziospecification}
E.~Bartocci, J.~Deshmukh, A.~Donz{\'e} \emph{et~al.}, ``Specification-based
  monitoring of cyber-physical systems: a survey on theory, tools and
  applications,'' in \emph{Lectures on Runtime Verification}, 2018, pp.
  135--175.

\bibitem{Adepu2016UsingPI}
\BIBentryALTinterwordspacing
S.~Adepu and A.~Mathur, ``{Using Process Invariants to Detect Cyber Attacks on
  a Water Treatment System},'' in \emph{{31st IFIP International Information
  Security and Privacy Conference (SEC)}}, vol. AICT-471, May 2016, pp.
  91--104. [Online]. Available: \url{https://hal.inria.fr/hal-01369545}
\BIBentrySTDinterwordspacing

\bibitem{attack_pcs}
A.~A. C{\'a}rdenas, S.~Amin, and Z.-S. Lin~et al., ``Attacks against process
  control systems: Risk assessment, detection, and response,'' in
  \emph{ASIACCS}, 2011, p.~12.

\bibitem{PumpRecall2019}
``Medtronic recalls minimed insulin pump models for cybersecurity risks,''
  \url{https://www.healio.com/news/endocrinology/20190627/medtronic-recalls-minimed-insulin-pump-models-for-cybersecurity-risks}.

\bibitem{PumpRecall2-2019}
``Fda issues class i recall of certain medtronic insulin pumps,''
  \url{https://beyondtype1.org/medtronic-recall/}.

\bibitem{MedicalCyberattack2019}
``Medical device cyber attacks: Tv plot or dangerous reality?''
  \url{https://www.drugwatch.com/news/2019/07/11/medical-device-cyber-attacks-tv-plot-or-dangerous-reality/}.

\bibitem{marcovecchio2017complications}
L.~Marcovecchio, ``Complications of acute and chronic hyperglycemia,'' \emph{US
  Endocrinology}, vol. 13 (1), 2017.

\bibitem{Mehmed5794}
A.~Mehmed, W.~Steiner, and A.~Causevic, ``Formal verification of an approach
  for systematic false positive mitigation in safe automated driving system,''
  April 2020.

\bibitem{fischer2011history}
H.~Fischer, \emph{A history of the central limit theorem: From classical to
  modern probability theory}.\hskip 1em plus 0.5em minus 0.4em\relax Springer,
  2011.

\bibitem{mode_adversarial_2020}
\BIBentryALTinterwordspacing
G.~R. Mode and K.~A. Hoque, ``Adversarial {Examples} in {Deep} {Learning} for
  {Multivariate} {Time} {Series} {Regression},'' in \emph{49th Annual IEEE
  Applied Imagery Pattern Recognition (AIPR) workshop 2020}, Sep. 2020.
  [Online]. Available: \url{http://arxiv.org/abs/2009.11911}
\BIBentrySTDinterwordspacing

\bibitem{xu2017feature}
W.~Xu, D.~Evans, and Y.~Qi, ``Feature squeezing: Detecting adversarial examples
  in deep neural networks,'' in \emph{Network and Distributed Systems Security
  Symposium (NDSS) 2018}, 2017.

\bibitem{guo_black-box_2021}
\BIBentryALTinterwordspacing
S.~Guo, J.~Zhao, X.~Li \emph{et~al.}, ``A {Black}-{Box} {Attack} {Method}
  against {Machine}-{Learning}-{Based} {Anomaly} {Network} {Flow} {Detection}
  {Models},'' \emph{Security and Communication Networks}, vol. 2021, Jan. 2021.
  [Online]. Available: \url{https://doi.org/10.1155/2021/5578335}
\BIBentrySTDinterwordspacing

\bibitem{zhou2018transferable}
W.~Zhou, X.~Hou, Y.~Chen \emph{et~al.}, ``Transferable adversarial
  perturbations,'' in \emph{Proceedings of the European Conference on Computer
  Vision (ECCV)}, 2018, pp. 452--467.

\bibitem{mode_crafting_2020}
\BIBentryALTinterwordspacing
G.~R. Mode and K.~A. Hoque, ``\BIBforeignlanguage{en}{Crafting {Adversarial}
  {Examples} for {Deep} {Learning} {Based} {Prognostics} ({Extended}
  {Version})},'' 2020. [Online]. Available:
  \url{https://arxiv.org/abs/2009.10149}
\BIBentrySTDinterwordspacing

\bibitem{openSourceOpenAPS}
``Principles of an open artificial pancreas system (openaps),''
  \url{https://openaps.org/reference-design}.

\bibitem{BBcontroller}
E.~Chertok~Shacham, H.~Kfir, N.~Schwartz \emph{et~al.}, ``Glycemic control with
  a basal-bolus insulin protocol in hospitalized diabetic patients treated with
  glucocorticoids: a retrospective cohort,'' \emph{BMC Endocr Disord}, vol.
  18(75), pp. 1472--6823, 2018.

\bibitem{openSourceGlucosym}
``Glucosym,'' \url{https://github.com/Perceptus/GlucoSym}.

\bibitem{man2014uva}
C.~D. Man, F.~Micheletto, D.~Lv \emph{et~al.}, ``The uva/padova type 1 diabetes
  simulator: new features,'' \emph{Journal of diabetes science and technology},
  vol.~8, no.~1, pp. 26--34, 2014.

\bibitem{zhang2018cost}
X.~Zhang and D.~Evans, ``Cost-sensitive robustness against adversarial
  examples,'' \emph{Seventh International Conference on Learning
  Representations}, 2018.

\bibitem{basak_universal_2021}
A.~Basak, P.~Rathore, S.~H. Nistala \emph{et~al.}, ``Universal {Adversarial}
  {Attack} on {Deep} {Learning} {Based} {Prognostics},'' in \emph{2021 20th
  {IEEE} {International} {Conference} on {Machine} {Learning} and
  {Applications} ({ICMLA})}, Dec. 2021, pp. 23--29.

\bibitem{DONG20217675}
\BIBentryALTinterwordspacing
W.~Dong, H.~Sun, J.~Tan \emph{et~al.}, ``Short-term regional wind power
  forecasting for small datasets with input data correction, hybrid neural
  network, and error analysis,'' \emph{Energy Reports}, vol.~7, pp. 7675--7692,
  2021. [Online]. Available:
  \url{https://www.sciencedirect.com/science/article/pii/S2352484721011665}
\BIBentrySTDinterwordspacing

\bibitem{lstm2015lipton}
\BIBentryALTinterwordspacing
Z.~C. Lipton, D.~C. Kale, C.~Elkan \emph{et~al.}, ``Learning to {Diagnose} with
  {LSTM} {Recurrent Neural Networks},'' 2015. [Online]. Available:
  \url{https://arxiv.org/abs/1511.03677}
\BIBentrySTDinterwordspacing

\bibitem{yasar_dsn2020}
M.~Yasar and H.~Alemzadeh, ``Real-time context-aware detection of unsafe events
  in robot-assisted surgery,'' in \emph{2020 50th Annual IEEE/IFIP
  International Conference on Dependable Systems and Networks (DSN)}, 2020, pp.
  385--397.

\bibitem{moon_ensemble_2022}
\BIBentryALTinterwordspacing
J.-H. Moon, J.-H. Yu, and K.-A. Sohn, ``\BIBforeignlanguage{en}{An ensemble
  approach to anomaly detection using high- and low-variance principal
  components},'' \emph{\BIBforeignlanguage{en}{Computers and Electrical
  Engineering}}, vol.~99, p. 107773, Apr. 2022. [Online]. Available:
  \url{https://www.sciencedirect.com/science/article/pii/S0045790622000714}
\BIBentrySTDinterwordspacing

\bibitem{rigaki_survey_2021-1}
\BIBentryALTinterwordspacing
M.~Rigaki and S.~Garcia, ``A {Survey} of {Privacy} {Attacks} in {Machine}
  {Learning},'' \emph{arXiv:2007.07646 [cs]}, Apr. 2021, arXiv: 2007.07646.
  [Online]. Available: \url{http://arxiv.org/abs/2007.07646}
\BIBentrySTDinterwordspacing

\bibitem{tramer2017ensemble}
F.~Tram{\`e}r, A.~Kurakin, N.~Papernot \emph{et~al.}, ``Ensemble adversarial
  training: Attacks and defenses,'' \emph{arXiv preprint arXiv:1705.07204},
  2017.

\bibitem{ganin2016domain}
Y.~Ganin, E.~Ustinova, H.~Ajakan \emph{et~al.}, ``Domain-adversarial training
  of neural networks,'' \emph{The journal of machine learning research},
  vol.~17, no.~1, pp. 2096--2030, 2016.

\bibitem{wong2020fast}
\BIBentryALTinterwordspacing
E.~Wong, L.~Rice, and J.~Z. Kolter, ``Fast is better than free: Revisiting
  adversarial training,'' in \emph{International Conference on Learning
  Representations}, 2020. [Online]. Available:
  \url{https://openreview.net/forum?id=BJx040EFvH}
\BIBentrySTDinterwordspacing

\bibitem{towards2016nicolas}
\BIBentryALTinterwordspacing
N.~Papernot, P.~McDaniel, A.~Sinha \emph{et~al.}, ``Towards the science of
  security and privacy in machine learning,'' 2016. [Online]. Available:
  \url{https://arxiv.org/abs/1611.03814}
\BIBentrySTDinterwordspacing

\bibitem{rifai2011contractive}
\BIBentryALTinterwordspacing
S.~Rifai, P.~Vincent, X.~Muller \emph{et~al.}, ``Contractive auto-encoders:
  Explicit invariance during feature extraction,'' in \emph{ICML}, 2011, pp.
  833--840. [Online]. Available:
  \url{https://icml.cc/2011/papers/455_icmlpaper.pdf}
\BIBentrySTDinterwordspacing

\bibitem{gu2014towards}
\BIBentryALTinterwordspacing
S.~Gu and L.~Rigazio, ``Towards deep neural network architectures robust to
  adversarial examples,'' in \emph{3rd International Conference on Learning
  Representations}, 2015. [Online]. Available:
  \url{http://arxiv.org/abs/1412.5068}
\BIBentrySTDinterwordspacing

\end{thebibliography}

    
    
    
    
    
    
\end{document}